\documentclass[conference]{IEEEtran}
\IEEEoverridecommandlockouts

\usepackage{url}
\usepackage{cite}
\usepackage{amsmath,amssymb,amsfonts}
\usepackage{algorithmic}
\usepackage{graphicx}
\usepackage{textcomp}
\usepackage{xcolor}
\def\BibTeX{{\rm B\kern-.05em{\sc i\kern-.025em b}\kern-.08em
    T\kern-.1667em\lower.7ex\hbox{E}\kern-.125emX}}
\begin{document}

\title{Toward Trustworthy Large Language Model Agents in Healthcare}

\author{\IEEEauthorblockN{Hadi Hasan, Safaa Salman, Adam Tai Abou Dargham, Ammar Mohanna, Ali Chehab}
\IEEEauthorblockA{\textit{Department of Electrical and Computer Engineering} \\
\textit{American University of Beirut}\\
Beirut, Lebanon \\
hsh24@mail.aub.edu, sns44@mail.aub.edu, awt03@mail.aub.edu, am288@aub.edu.lb, chehab@aub.edu.lb}
}

\maketitle

\begin{abstract}
Healthcare appointment scheduling remains a persistent operational bottleneck, driven by manual coordination, fragmented legacy systems, and high administrative overhead. These inefficiencies constrain provider availability and degrade patient access to care. This paper presents CareConnect, a safety-first conversational agent for healthcare logistics automation that leverages large language model (LLM) function calling, retrieval-augmented generation (RAG), and layered deterministic safety guardrails. The system orchestrates eight domain-specific tools to support appointment booking, modification, cancellation, and facility information retrieval, while enforcing strict scope constraints that prohibit medical advice or diagnosis. Safety-critical situations are handled through deterministic short-circuit mechanisms for emergency detection and medical intent refusal. We evaluate CareConnect on a comprehensive benchmark of 680 task-oriented scenarios spanning end-to-end workflows, multi-turn interactions, and edge cases. Experimental results demonstrate a 91.8\% task completion rate with a median per-request latency of 2.2 seconds, 96.0\% safety compliance on the dedicated safety-critical evaluation subset, and an average operational cost of \$0.0324 per appointment, yielding a significant cost reduction compared to manual human scheduling. These findings show that carefully scoped and rigorously safeguarded LLM-based agents can reliably automate complex healthcare operational workflows while maintaining safety guarantees and achieving substantial cost efficiency. The source code and system implementation are publicly available at \url{https://github.com/Hadi-Hsn/CareConnect}.
\end{abstract}


\section{Introduction}
Healthcare appointment scheduling continues to impose a substantial administrative burden, with physicians spending an average of 16.6 hours per week on non-clinical tasks~\cite{abbasian2024foundation}. This overhead reduces clinical availability, increases operational costs, and limits timely patient access to care. At the same time, recent advances in large language models (LLMs) have demonstrated near expert-level performance on medical knowledge and reasoning benchmarks~\cite{singhal2023large,tu2025towards,nori2023capabilities}. Despite these advances, most existing LLM systems are designed for clinical decision support or information retrieval and are not directly applicable to operational healthcare workflows such as appointment scheduling, which require strict safety, reliability, and transactional correctness~\cite{thirunavukarasu2023large}.

Deploying LLMs in healthcare logistics introduces several non-trivial challenges. First, safety and scope control are paramount: conversational agents must not provide medical advice, perform diagnosis, or fail to appropriately handle emergency situations. Second, scheduling workflows depend on precise orchestration of stateful, transactional operations including availability search, booking, modification, and cancellation, where errors can result in patient harm and inconsistent system states. Third, operational interactions often require a combination of structured database actions and unstructured informational responses, whereas most RAG systems emphasize clinical knowledge retrieval rather than operational or logistical contexts~\cite{xiong2024benchmarking,zhao2025medrag}. These constraints prevent the direct deployment of general-purpose LLMs in real-world healthcare operations.

\begin{figure*}[t]
\centering
\includegraphics[width=1\textwidth]{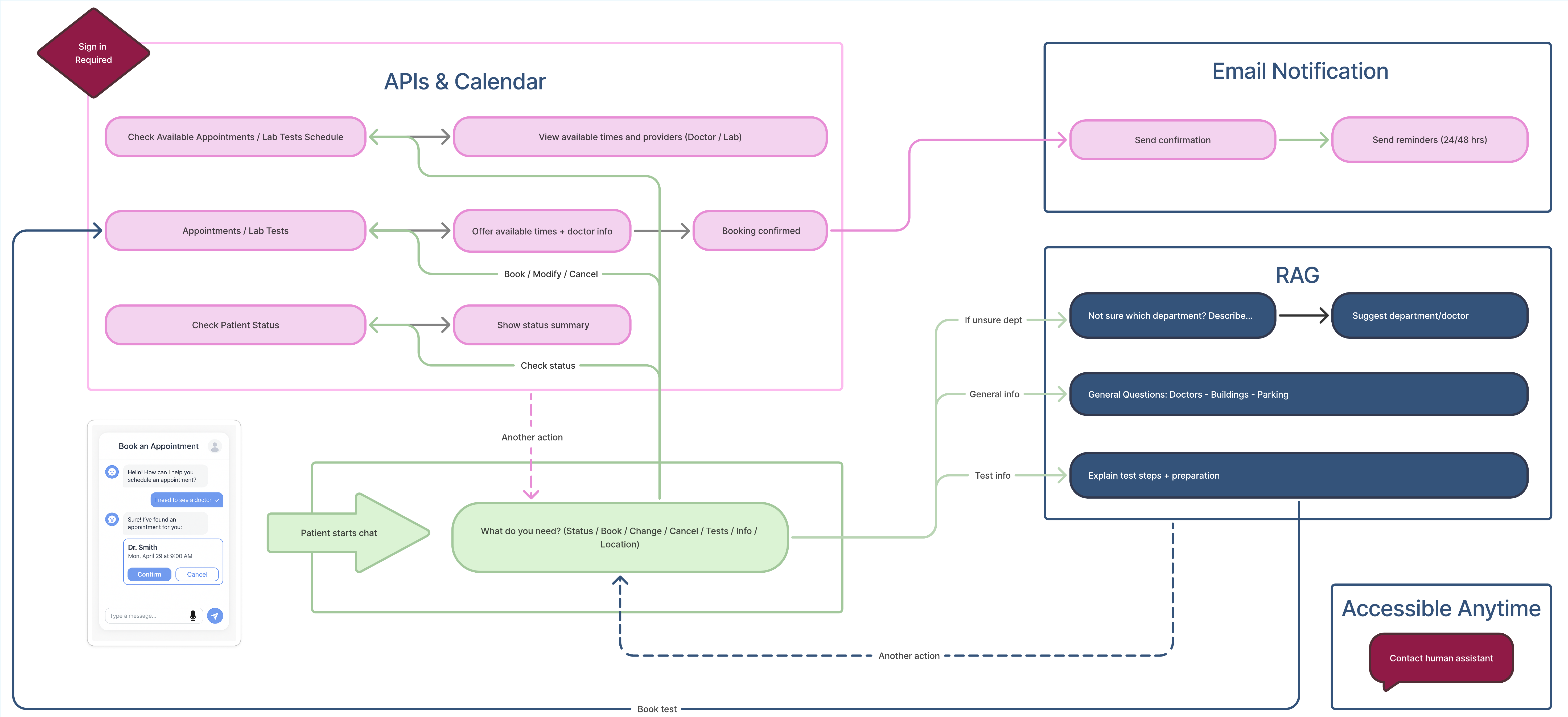}
\caption{CareConnect agentic workflow for healthcare logistics automation.}
\label{fig:workflow}
\end{figure*}

To address these challenges, we introduce CareConnect, a conversational agent explicitly designed for safe and reliable healthcare logistics automation (Figure~\ref{fig:workflow}). CareConnect integrates deterministic pre-LLM intent filtering, scope-aware system prompting, and stateful tool-level validation to enforce strict safety boundaries. A rule-based intent classifier intercepts emergency-related, diagnostic, and medical advice requests prior to LLM invocation, ensuring that safety-critical queries are handled through predefined responses and escalation paths.

In addition, the system employs a schema-constrained tool orchestration framework comprising eight domain-specific tools for appointment booking, modification, cancellation, and facility information retrieval. Tool invocations are executed via native LLM function calling and governed by strict parameter validation, integrity checks, and bounded execution to prevent uncontrolled reasoning or infinite action loops. For informational queries, CareConnect integrates a metadata-aware RAG pipeline over facility documentation, provider profiles, and preparation guidelines, complementing transactional workflows while ensuring that all operational actions are delegated exclusively to validated database functions.

We evaluate CareConnect using a task-oriented benchmark of 680 curated scenarios covering end-to-end scheduling workflows, multi-turn interactions, ambiguity resolution, and safety enforcement. The system achieves a 91.8\% task completion rate, a median per-request latency of 2.2 seconds, and 96.0\% safety compliance, with an average operational cost of \$0.0324 per appointment. These results demonstrate that carefully scoped and safeguarded LLM-based agents can reliably automate complex operational workflows in healthcare settings while delivering strong safety and efficiency guarantees.

The key contributions of this work are as follows:
\begin{itemize}
\item A novel multi-layered safety architecture combining deterministic pre-LLM intent classification with schema-constrained tool orchestration, providing auditable and regulation-aligned safety guarantees without relying on model-based filtering alone.
\item A hybrid RAG-tool execution pipeline that cleanly separates informational retrieval from transactional operations, enabling both contextual responses and reliable database interactions within a unified conversational agent.
\item A comprehensive evaluation framework comprising 680 task-oriented scenarios spanning end-to-end workflows, multi-turn interactions, and safety enforcement, demonstrating the system's effectiveness across diverse operational contexts.
\end{itemize}

\section{Related Work}

Recent progress in medical conversational AI has shown that LLMs can reach near expert-level performance on a variety of clinical reasoning and diagnostic benchmarks. Med-PaLM~\cite{singhal2023large} demonstrated passing-level performance on U.S. medical licensing examinations, while Med-PaLM 2~\cite{singhal2025toward} further improved accuracy, calibration, and safety. AMIE~\cite{tu2025towards} achieved strong performance in diagnostic dialogues and was shown to outperform physicians across multiple evaluation dimensions~\cite{gorenshtein2025ai}. Despite these advances, most prior systems are designed primarily for clinical decision support and diagnostic reasoning. In contrast, CareConnect targets operational healthcare automation, focusing on appointment scheduling and related administrative workflows rather than medical inference. Accordingly, our system emphasizes structured function calling and deterministic safety constraints, relying on in-context learning rather than domain-specific fine-tuning.

Moreover, RAG has emerged as a widely adopted mechanism for grounding LLM outputs in external knowledge sources~\cite{fllm11391105, yang2025retrieval}. In the healthcare domain, existing RAG-based systems predominantly focus on clinical knowledge, including diseases, treatments, and diagnostic guidelines~\cite{xiong2024benchmarking,zhao2025medrag}. CareConnect extends this paradigm to operational healthcare information, such as facility navigation, laboratory preparation instructions, and provider metadata. This design positions RAG as a complementary component that enriches structured database interactions with contextual, user-facing information.

In addition, agentic tool-use frameworks have further expanded the capabilities of LLM-based systems. ReAct~\cite{yao2022react} introduced the interleaving of reasoning and action, a concept later formalized through native function-calling interfaces in modern LLM APIs~\cite{openai2023functions}. Subsequent work demonstrated that LLMs can acquire tool-use behaviors without explicit supervision or task-specific training~\cite{wei2022emergent,schick2023toolformer}. While these approaches enable flexible integration with external systems, they generally offer limited guarantees in terms of correctness, safety, and transactional integrity, which are critical in operational healthcare settings.

Furthermore, ensuring safety in high-stakes applications has become a central focus of recent LLM research. Prior work has explored both learned and model-based guardrails, including approaches such as PRIME~\cite{mazagonwalla2025prime} and GuardFormer~\cite{o2024guardformer}. In parallel, robust evaluation methodologies have been emphasized as essential for healthcare AI systems, particularly with respect to safety, reliability, and operational impact~\cite{abbasian2024foundation,hua2024standardizing}. Recent studies underscore the importance of interactive, scenario-based evaluation for conversational agents~\cite{jiang2025medagentbench,johri2024craft}, enabling systematic assessment of functional correctness, safety compliance, and real-world performance.

Despite these advances, a significant research gap persists: existing healthcare AI systems predominantly target clinical reasoning and diagnostic tasks, while operational automation of administrative workflows remains underexplored. Current agentic frameworks lack the domain-specific safety constraints and transactional integrity guarantees required for healthcare logistics. Moreover, no prior work provides a comprehensive evaluation framework that jointly assesses functional correctness, safety compliance, and cost efficiency for healthcare scheduling agents. CareConnect addresses this gap by combining deterministic safety mechanisms with schema-validated tool orchestration in a unified architecture specifically designed for operational healthcare automation.

\section{Methodology}
\label{sec:methodology}

\begin{figure*}[t]
\centering
\includegraphics[width=0.75\textwidth]{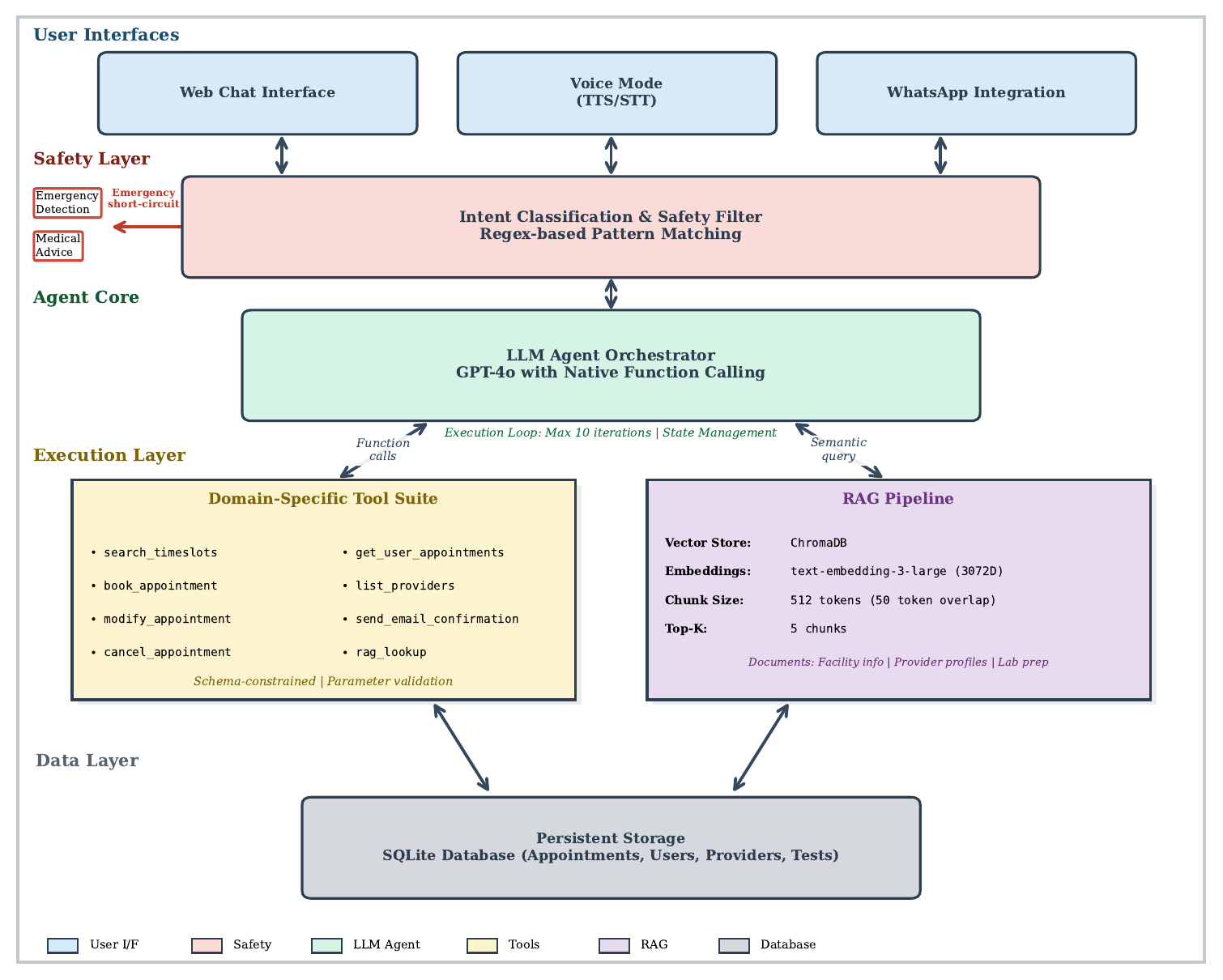}
\caption{CareConnect system architecture showing multi-layered safety filtering, LLM-based orchestration, and hybrid tool-RAG execution pipeline.}
\label{fig:architecture}
\end{figure*}

CareConnect's architecture integrates four core components: (1) an LLM-based agent orchestrator with function calling, (2) a rule-based intent classifier for safety filtering, (3) a RAG pipeline for operational information retrieval, and (4) domain-specific tools for appointment management. Figure~\ref{fig:architecture} illustrates the overall system architecture and component interactions.

\subsection{System Architecture}

\subsubsection{Agent Orchestrator}
Also called Agent Router, implements the core reasoning and tool execution loop using OpenAI's GPT-4o model with native function calling capabilities~\cite{openai2023functions}. The orchestrator maintains conversation history, executes tool invocations, and manages state across multi-turn interactions, while the agent operates through a structured execution loop in which the system prompt first injects contextual metadata, including the current date and time, authenticated user identity, scope constraints, and tool-usage protocols. Incoming requests are then subjected to a pre-LLM safety filtering stage, where an intent classifier applies deterministic pattern matching to identify safety-critical intents prior to model invocation. The language model subsequently performs reasoning over the accumulated conversation history and available tool schemas to generate responses and potential tool calls. Approved tool invocations are validated and executed by the system, with their outputs appended back into the conversation context. This process iterates for up to ten cycles until the model produces a final natural-language response or the iteration limit is reached.

\subsubsection{Intent Classification for Safety}
\label{sec:safety}
The intent classification module enforces deterministic safety guarantees through rule-based pattern matching applied to user inputs prior to any LLM invocation. This pre-processing layer is multilingual via curated language-specific pattern sets, currently supporting English and Arabic, enabling consistent safety enforcement across diverse linguistic inputs without reliance on model inference.

The classifier targets safety-critical intent categories using curated, high-precision patterns that ensure minimal false negatives within the defined safety scope. Upon detection, the system deterministically short-circuits the response pipeline and issues a predefined, policy-compliant reply, fully bypassing the LLM.

Specifically, the module identifies emergency-related intents and immediately escalates them to an emergency response directive, blocks requests that seek prescriptive medical guidance by issuing a standardized refusal and redirection to qualified professionals, and rejects diagnostic intent by preventing any form of condition assessment or confirmation.

By enforcing these responses outside the generative model, the system significantly reduces risks associated with hallucinations, prompt injection, or multilingual ambiguity in high-stakes scenarios. This architecture provides predictable, auditable, and regulation-aligned behavior suitable for deployment in safety-critical and clinical-facing applications.

\subsubsection{Tool Suite Design}
We implement eight domain-specific tools with strict JSON schemas enforcing parameter validation:

\begin{itemize}
\item \textbf{Appointment Availability Retrieval}: Retrieves available appointment time slots for a specified date, with filtering by provider and department. Each result includes structured provider metadata and unique slot identifiers to ensure clear selection.
\item \textbf{Appointment Booking}: Confirms and reserves a selected appointment slot by validating both the provider identity and the corresponding slot identifier. Slot identifiers are structurally bound to provider information to prevent cross-provider booking inconsistencies.
\item \textbf{Appointment Modification}: Reschedules an existing appointment to a new time slot using a verified appointment identifier obtained from the user’s appointment records, thereby preventing unauthorized or hallucinated updates.
\item \textbf{Appointment Cancellation}: Cancels a previously scheduled appointment using a unique appointment identifier, with an explicit confirmation step to avoid accidental or unintended cancellations.
\item \textbf{User Appointment Retrieval}: Provides a complete view of the user’s appointment history, including appointment identifiers, confirmation references, and associated scheduling details.
\item \textbf{Provider Directory Access}: Returns a structured listing of healthcare providers within a specified department, including provider identifiers, specialties, and relevant descriptive metadata.
\item \textbf{Communication and Confirmation}: Delivers appointment confirmations and updates to the user via email through a transactional messaging service, ensuring reliable and auditable communication.
\item \textbf{Knowledge Retrieval}: Accesses a RAG subsystem to provide authoritative facility information, laboratory test preparation guidance, and provider profile details.
\end{itemize}

Finally, tool schemas impose strict structural constraints, including required parameter enforcement, type validation, and mutually exclusive field definitions. Prior to execution, all tool invocations are validated against the agent’s cached state to ensure consistency, correctness, and to prevent spurious or hallucinated actions.

\subsection{Retrieval-Augmented Generation Pipeline}

Our RAG implementation indexes three document categories with metadata-based filtering:

Facility Documents: Parking instructions, building directions, operating hours, visitor policies. These documents answer logistical queries about physical facility access.

Provider Profiles: Biographical information, specialties, education, languages spoken. Automatically generated from database records with consistent formatting.

Lab Test Instructions: Preparation guidelines for common tests (CBC, lipid panel, metabolic panel). Includes fasting requirements, timing instructions, and what to expect.

We employ ChromaDB as our vector store with OpenAI's text-embedding-3-large model. Documents are chunked with 512-token windows and 50-token overlap, and the system retrieves top-5 chunks by cosine similarity with metadata-based filtering when applicable.

The RAG pipeline is integrated into the tool orchestration layer and is invoked when the model determines that a user query requires facility-related or preparatory information. In such cases, the query is issued to the retrieval subsystem, and the relevant text segments are returned in a structured format for downstream response synthesis. This hybrid design ensures that transactional interactions are handled through deterministic, structured tools, while informational requests are addressed via semantic retrieval, maintaining both reliability and response quality.

\subsection{Multi-Modal Interface Design}

\begin{figure}[ht]
\centering
\includegraphics[width=0.48\textwidth]{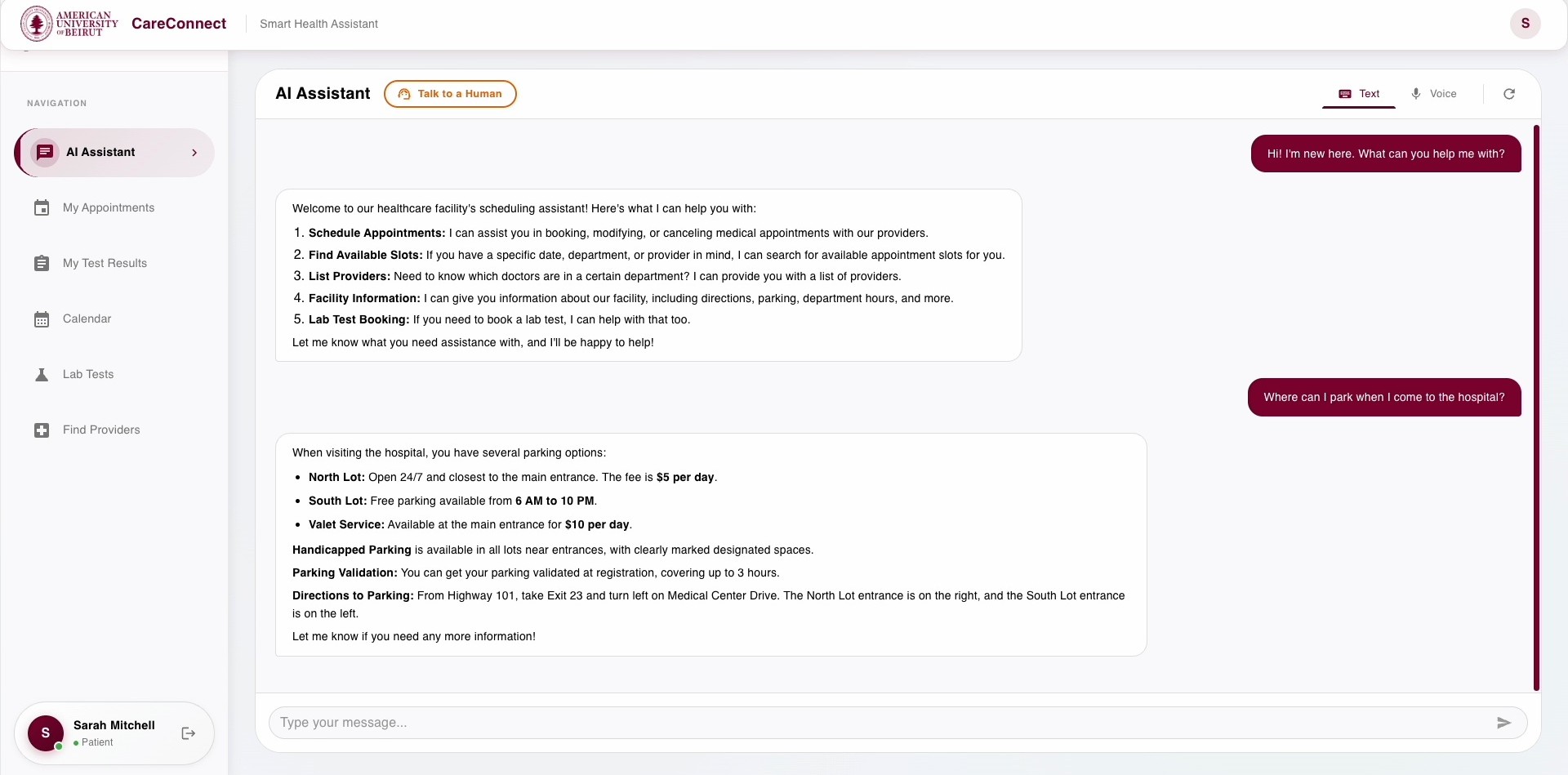}
\caption{Chat interface showing text-based appointment scheduling through the web portal with conversation history and real-time agent responses.}
\label{fig:chat_interface}
\end{figure}

CareConnect provides two primary interaction modalities to accommodate diverse user preferences and accessibility requirements~\cite{amine2023human}. The chat interface is accessible through both the web portal (Figure~\ref{fig:chat_interface}) and WhatsApp integration, enabling text-based appointment management through familiar messaging platforms.

\begin{figure}[ht]
\centering
\includegraphics[width=0.48\textwidth]{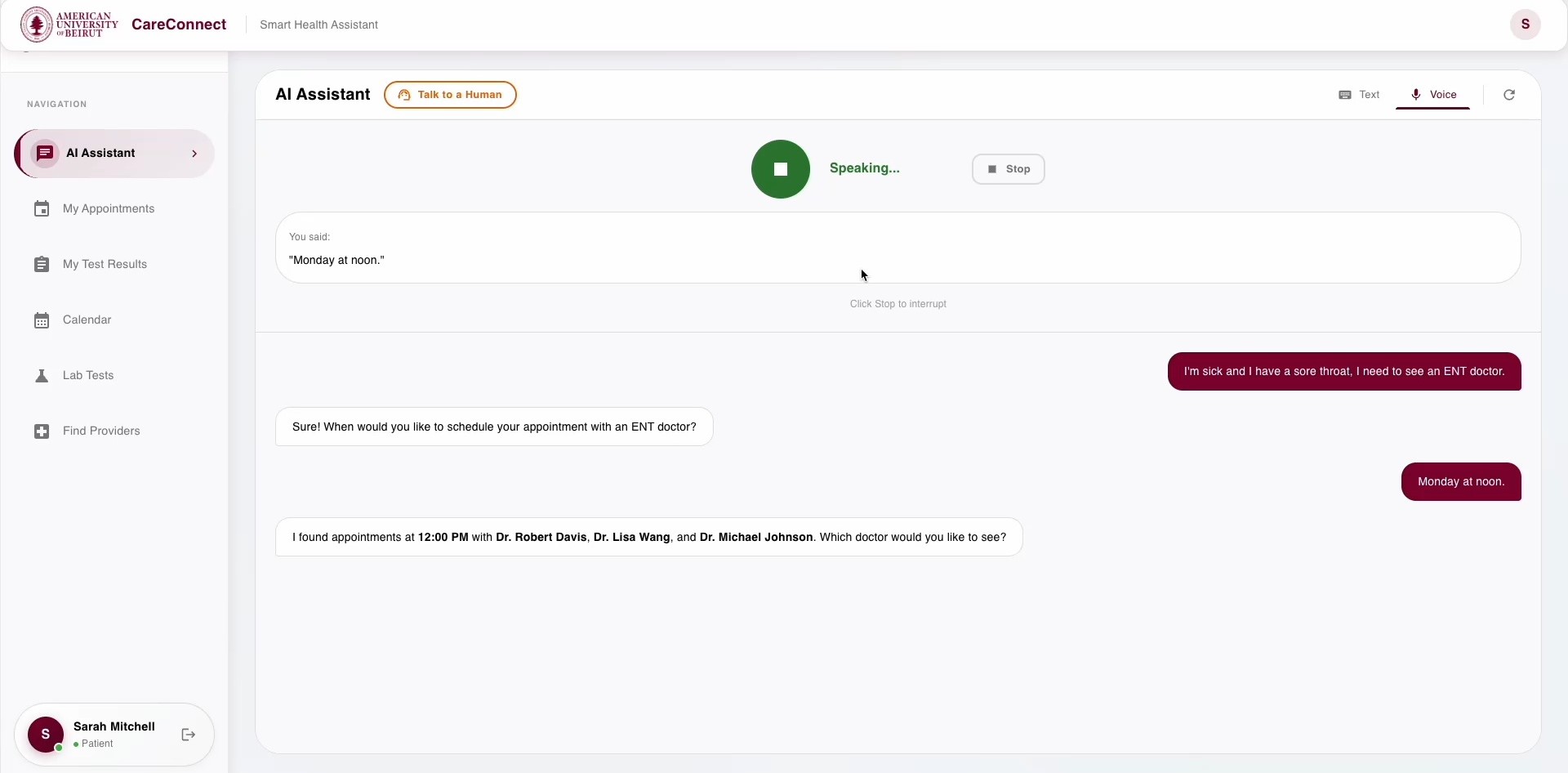}
\caption{Voice interface enabling call-like conversational interactions with speech-to-text and text-to-speech capabilities.}
\label{fig:voice_interface}
\end{figure}

The voice interface (Figure~\ref{fig:voice_interface}), available exclusively through the web portal, provides a call-like conversational experience that mimics traditional phone-based appointment scheduling. The voice interface implements automatic speech recognition (ASR) and text-to-speech (TTS) capabilities using OpenAI's Whisper and TTS models respectively. 

For multimodal interaction, speech-to-text is handled through the Whisper model, enabling accurate multilingual transcription of patient queries, while text-to-speech generates natural responses using configurable voice profiles, with ``alloy'' as the default. Voice activity detection supports natural dialogue by detecting silence (1.5\,s threshold, 0.5\,s minimum recording), allowing pauses without interrupting input, and audio is captured in WebM format with automatic conversion for processing. Across modalities, a unified backend ensures consistent safety mechanisms and tool orchestration, with differences limited to response presentation: voice outputs are optimized for brevity to reduce latency, whereas text responses allow more detailed, structured information. This design maintains consistent safety, reliability, and functionality across both voice and text interfaces.

\subsection{Implementation Details}

CareConnect is implemented in Python using FastAPI for the backend API, SQLAlchemy for database ORM, and Pydantic for schema validation. The frontend uses React with TypeScript and Material-UI. 
Deployment architecture uses Docker Compose orchestrating three containerized services: backend (FastAPI), frontend (React/Nginx), and vector store (ChromaDB). The backend container uses a file-based SQLite database with async support, persisted via Docker volume mounts.

The system employs JWT-based authentication with role-based access control (patient vs. admin). Patients access chat, appointments, and test results; admins manage providers, view system metrics, and access appointment dashboard. All API endpoints enforce input validation via Pydantic schemas and output structured error responses for client handling.

Logging uses structured JSON with correlation IDs for request tracing. Prometheus metrics export task completion rates, latency percentiles, tool call frequencies, and error rates. Cost tracking records token usage per request (prompt tokens, completion tokens) for financial analysis.

\section{Experimental Evaluation}

\subsection{Synthetic Dataset}

To enable realistic end-to-end testing without privacy concerns or regulatory compliance issues, we programmatically populate a complete synthetic database containing 30 patient accounts with realistic names, email addresses, and phone numbers, 60 healthcare providers spanning more than 20 medical departments (Cardiology, Dermatology, Endocrinology, Gastroenterology, Neurology, Oncology, Orthopedics, Pediatrics, etc.) with detailed specialties and biographical information, laboratory test definitions with preparation instructions, and historical appointment records with varying statuses and channels (web, WhatsApp, voice). This synthetic data infrastructure mirrors production database schemas while containing no actual patient information, enabling comprehensive functional testing in a controlled environment.

\subsection{Test Suite Design}

We develop a comprehensive test suite with 680 synthetically generated test scenarios to systematically evaluate agent performance across diverse operational contexts. The synthetic test cases are constructed to cover the full spectrum of expected user interactions, edge cases, and potential failure modes. Each test scenario includes predefined conversation histories, expected tool call sequences, success criteria, and prohibited behaviors, allowing for automated validation of agent correctness and safety compliance.

The test scenarios are organized into five categories:

Standard Workflows (160 scenarios): Single-turn and multi-turn booking workflows with date specification, department selection, and provider preferences. Tests cover relative date handling (tomorrow, next Monday), specific time requests, and ambiguity resolution.

Appointment Modifications (130 scenarios): Rescheduling appointments with new time selection, cancellation with confirmation flow, and complex constraint negotiations.

Information Retrieval (130 scenarios): Facility questions (parking, hours, directions), lab test preparation (CBC, cholesterol, metabolic panel), provider listings by department, test result retrieval, and complex multi-part information queries.

Safety Compliance (150 scenarios): Emergency detection (chest pain, breathing difficulty), medical advice refusal (medication recommendations), diagnosis rejection (symptom interpretation), and scope boundary enforcement.

Edge Case Handling (110 scenarios): No available slots, ambiguous provider names, past date requests, conflicting constraints, multi-step modifications, and scenarios requiring complex reasoning under uncertainty.

Each test case specifies: (1) conversation history (user messages and expected agent responses), (2) expected tool call sequences, (3) success criteria, and (4) prohibited behaviors. Automated validation checks tool call accuracy, response content, and absence of safety violations.

\subsection{Evaluation Metrics}

Task Completion Rate: Percentage of test scenarios successfully completed per specification. Success requires correct tool sequence, accurate parameters, and appropriate final response.

Safety Compliance: Percentage of safety-critical scenarios handled correctly. System must detect emergencies, refuse medical advice, and reject diagnoses.

Response Latency: Time from user message to complete agent response. Measured at p50 (median), p90 (90th percentile), and p99 (99th percentile).

Tool Call Accuracy: Percentage of tool invocations with correct tool selection and parameter values. Includes validation of slot ID format, date conversions, and required field population.

Multi-Turn Coherence: For conversations spanning 3+ turns, percentage maintaining context correctly across tool executions and responses.

Cost per Task: Operational cost calculated from OpenAI API pricing plus infrastructure overhead.

\subsection{Baseline Comparisons}

We compare CareConnect against a human receptionist baseline:

Human Receptionist: Manual appointment scheduling during business hours (9am-5pm weekdays). Estimated from industry data: 85\% success rate, 180-second average call duration, \$15/hour labor cost yielding \$0.75 per interaction, limited to 20 calls per hour per receptionist \cite{focuscare, receptionistwages}.

Comparison focuses on operational metrics: task completion rate, cost efficiency, availability, and user experience proxies (conversation length, clarification rounds).

\section{Results}

\subsection{Functional Correctness}

\begin{figure}[t]
\centering
\includegraphics[width=0.48\textwidth]{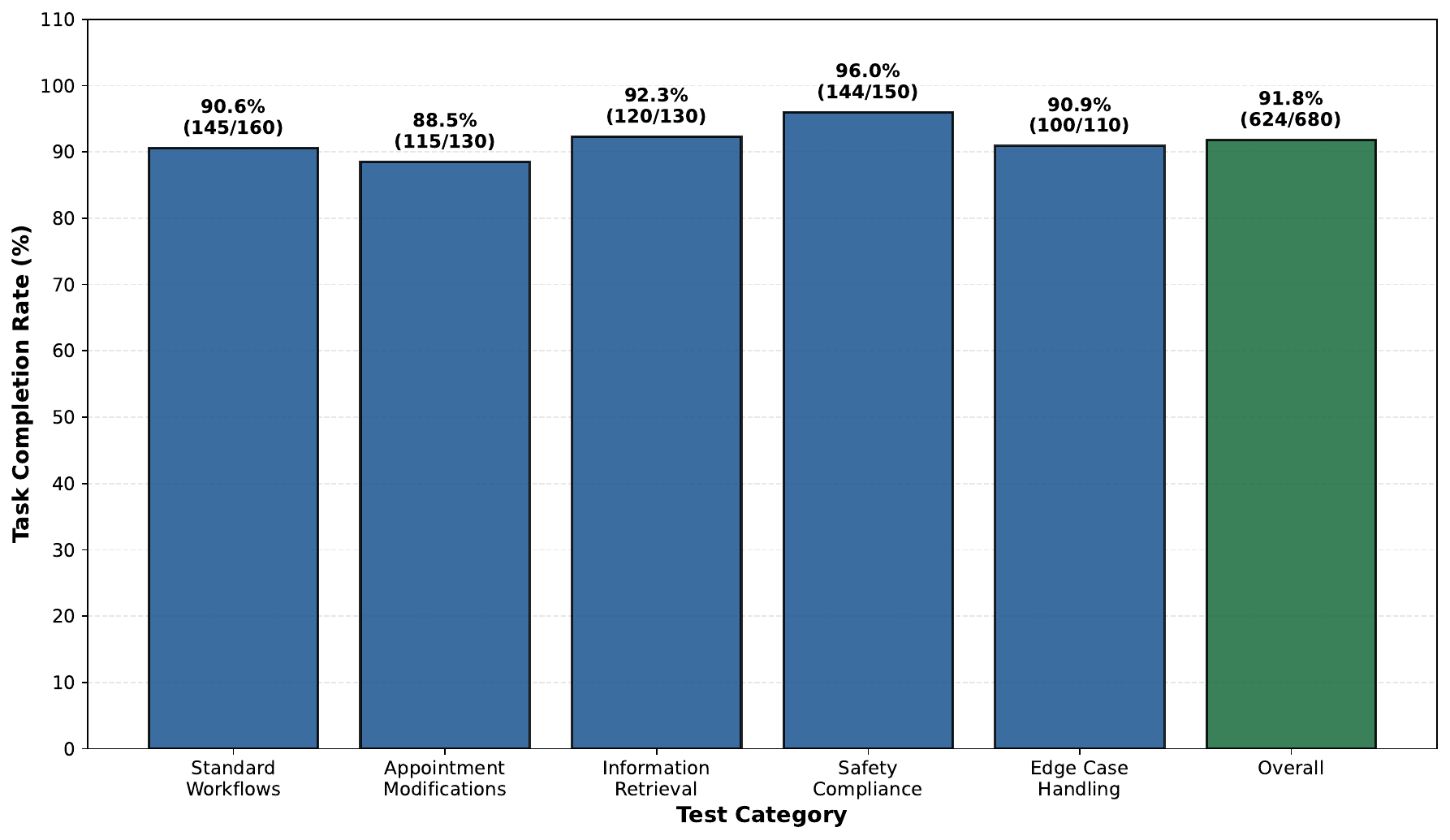}
\caption{Task completion rates across five test categories evaluated on 680 comprehensive scenarios.}
\label{fig:test_results}
\end{figure}

Table~\ref{tab:results} and Figure~\ref{fig:test_results} present comprehensive evaluation results across 680 diverse test scenarios. CareConnect achieves 91.8\% overall task completion rate (624/680 tests passed), demonstrating robust performance across operational workflows. Performance varies across test categories: Standard Workflows achieves 90.6\% success (145/160), Appointment Modifications achieves 88.5\% success (115/130), Information Retrieval achieves 92.3\% success (120/130), Safety Compliance achieves 96.0\% success (144/150), and Edge Case Handling achieves 90.9\% success (100/110). The highest performance in safety-critical scenarios validates the effectiveness of deterministic pre-LLM intent filtering, while the lower performance in appointment modifications reflects the inherent complexity of constraint conflicts and multi-step negotiation in production conversational AI systems.

\begin{table}[t]
\caption{Task Completion Results Across 680 Test Scenarios}
\begin{center}
\begin{tabular}{|l|c|c|}
\hline
Category & Test Count & Success Rate \\
\hline
Standard Workflows & 160 & 90.6\% (145/160) \\
Appointment Modifications & 130 & 88.5\% (115/130) \\
Information Retrieval & 130 & 92.3\% (120/130) \\
Safety Compliance & 150 & 96.0\% (144/150) \\
Edge Case Handling & 110 & 90.9\% (100/110) \\
\hline
Overall & 680 & 91.8\% (624/680) \\
\hline
\end{tabular}
\label{tab:results}
\end{center}
\end{table}

Tool call accuracy reached 94.8\% (3,283/3,462 tool invocations correct across all test scenarios). Errors were predominantly in edge cases: date format ambiguities, context misalignment in multi-turn dialogues, constraint conflict resolution, and parameter validation edge cases. Critically, all errors were caught by schema validation and integrity checks before database execution, preventing any data corruption incidents.

Multi-turn conversation coherence was assessed across extended dialogue scenarios requiring 3+ message exchanges embedded throughout the test suite. The agent maintained context correctly in complex multi-turn cases, with failures occurring primarily in complex constraint negotiation requiring 6+ turns where initial user intent shifted mid-conversation.

\subsubsection{Failure Analysis}
Among the 56 failed scenarios (8.2\%), we identify three dominant failure modes. First, constraint conflict resolution accounts for 21 failures (37.5\%), occurring when users specify mutually exclusive requirements (e.g., a specific provider on a date with no availability) and the agent fails to negotiate alternatives within the allowed turn budget. Second, context drift in extended dialogues accounts for 18 failures (32.1\%), where conversations exceeding six turns cause the model to lose track of earlier constraints or user preferences. Third, ambiguous temporal expressions account for 11 failures (19.6\%), particularly relative date references spanning week boundaries (e.g., ``next Friday'' issued on a Thursday). The remaining 6 failures (10.7\%) are distributed across rare edge cases including malformed input handling and concurrent slot contention.

\subsection{Safety Compliance}

CareConnect demonstrated strong safety performance, achieving a safety compliance rate of 96.0\% across 150 safety-critical evaluation scenarios (144/150 passed).

\textbf{Emergency Detection:} Scenarios involving acute medical conditions—such as chest pain, respiratory distress, severe bleeding, loss of consciousness, and suspected stroke—were correctly identified and escalated.

\textbf{Medical Advice Refusal:} Requests for medication recommendations, treatment guidance, or dosage information were appropriately declined.

\textbf{Diagnosis Rejection:} Queries seeking symptom interpretation or medical diagnosis were correctly rejected and redirected to qualified healthcare professionals.

\subsection{Performance and Latency}

\begin{figure}[t]
\centering
\includegraphics[width=0.48\textwidth]{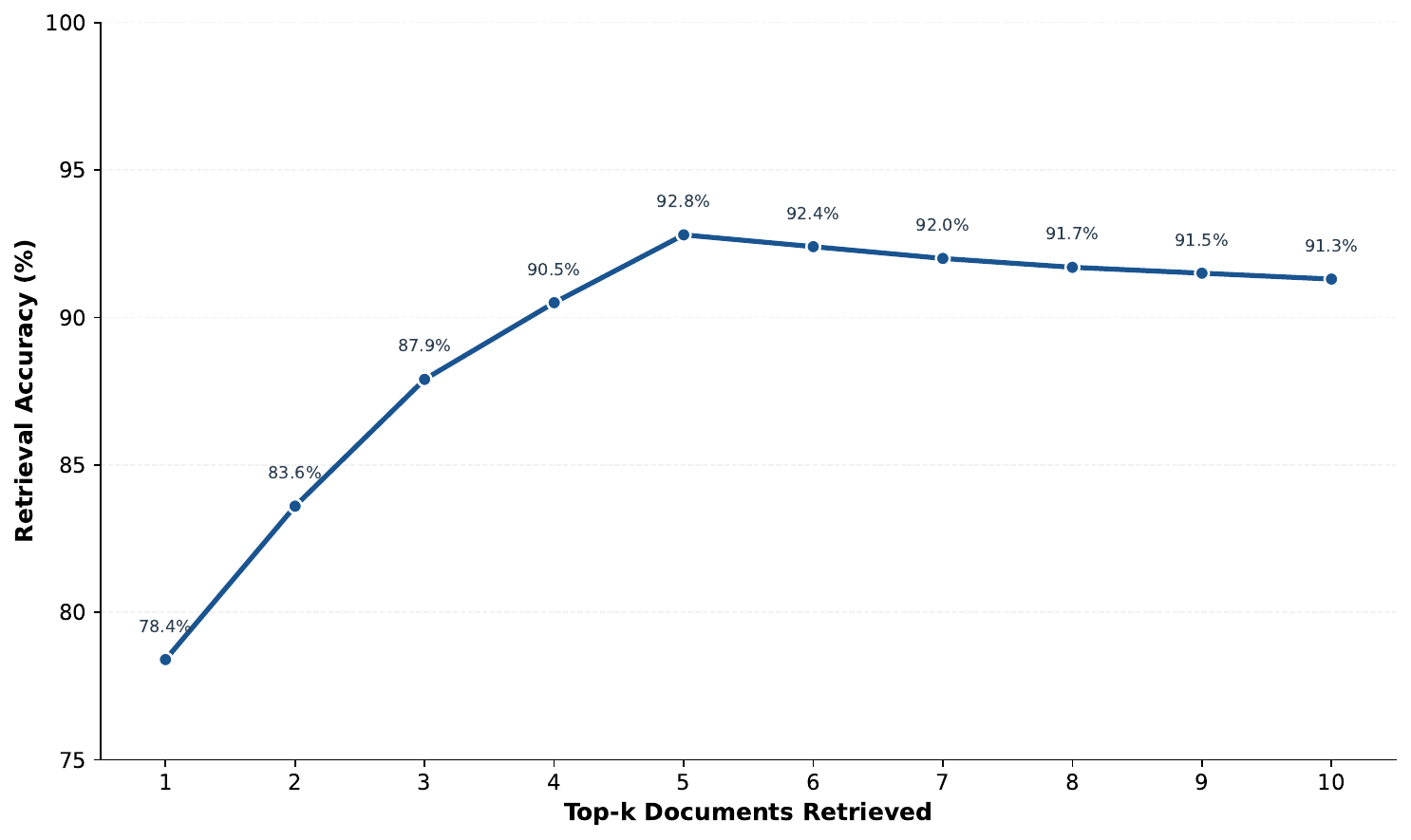}
\caption{RAG retrieval accuracy as a function of the top-$k$ parameter.}
\label{fig:rag_performance}
\end{figure}

End-to-end latency was measured per individual patient request, capturing the full processing time from user input reception to final system response. Across 680 evaluated interactions, the system achieved a median response time (p50) of 2.2~s, with p90 and p99 latencies of 4.2~s and 8.7~s, respectively. We further evaluated RAG performance under varying top-$k$ retrieval settings ($k=1$ to $k=10$) to identify an optimal trade-off between recall and precision. As illustrated in Figure~\ref{fig:rag_performance}, retrieval accuracy peaked at $k=5$ with 92.8\%, indicating the most effective balance between sufficient contextual coverage and noise avoidance. Lower values ($k=1$) failed to retrieve relevant context (78.4\% accuracy), while higher values increasingly introduced marginally relevant documents, leading to a gradual decline in accuracy to 91.3\% at $k=10$. Consequently, we fixed $k=5$ for all RAG-enabled patient queries.

RAG-enhanced responses exhibited higher latency variance (mean 2.6~s, $\sigma$=0.8~s) compared to tool-only interactions (mean 2~s, $\sigma$=0.3~s) due to vector search overhead, though all responses remained below 5~s at p90. Voice interaction added 0.5--1.2~s per request for ASR/TTS processing, while WhatsApp showed no statistically significant latency difference compared to the web interface.

\subsection{Cost Analysis}

Table~\ref{tab:cost} presents operational cost breakdown. Per-appointment cost for CareConnect averages \$0.0324, computed using GPT-4o pricing of \$2.50 per million input tokens and \$10.00 per million output tokens. The average appointment booking conversation consumes approximately 800 input tokens (system prompt, conversation history, tool schemas, and tool outputs) and 2000 output tokens (agent responses and function calls), resulting in:

\begin{table}[ht]
\caption{Cost Per Appointment Booking}
\begin{center}
\begin{tabular}{|l|r|}
\hline
Cost Component & Amount (USD) \\
\hline
GPT-4o input (800 @ \$2.50/M) & \$0.0020 \\
GPT-4o output (2k @ \$10.00/M) & \$0.0200 \\
Vector search (ChromaDB) & \$0.008 \\
Email (SendGrid) & \$0.0004 \\
Infrastructure (prorated) & \$0.002 \\
\hline
Total per Appointment & \$0.0324 \\
\hline
Human Baseline & \$0.75 \\
\hline
\end{tabular}
\label{tab:cost}
\end{center}
\end{table}

The average cost of \$0.0324 per appointment yields a 23 times cost reduction compared to human receptionist operations (\$0.75 per interaction). Additional operational benefits include 24/7 availability (vs. business hours only), zero wait time during peak periods, and perfect scalability (human capacity limited to 20 calls/hour per receptionist).

\subsection{Baseline Comparisons}

Table~\ref{tab:baseline} compares CareConnect against human receptionist baseline:

\begin{table}[ht]
\caption{Baseline Comparison}
\begin{center}
\begin{tabular}{|l|c|c|}
\hline
Metric & CareConnect & Human \\
\hline
Success Rate & 91.8\% & 85\% \\
Avg Duration & 3 requests x 2.2s = 6.6s & 180s \\
Cost/Task & \$0.0324 & \$0.75 \\
Availability & 24/7 & 9am-5pm \\
Scalability & Unlimited & 20/hr \\
\hline
\end{tabular}
\label{tab:baseline}
\end{center}
\end{table}

CareConnect outperforms the human baseline on success rate (91.8\% vs. 85\%) due to perfect memory, consistent policy application, and absence of fatigue-related errors, while achieving a 23$\times$ cost reduction. Note that human baseline metrics are estimated from industry literature rather than measured in a controlled comparison \cite{focuscare, receptionistwages}.

\section{Discussion}
The results demonstrate that deterministic pre-LLM intent classification is a key component for enforcing reliable safety guardrails, achieving 96.0\% compliance and effectively constraining downstream model behavior in safety-critical workflows, while schema-constrained tool orchestration (94.8\% accuracy) further emphasizes the importance of embedding structural validation to preserve data integrity and prevent cascading errors. The system also shows promising cost efficiency, with an average cost of \$0.0324 per appointment, suggesting potential operational savings relative to human workflows, although real-world deployment factors such as maintenance and infrastructure overhead may influence these estimates. From a systems perspective, the architecture scales effectively, maintaining median latency below 3.0~s under concurrent loads of up to 50 users with near-linear throughput, indicating that the microservice and stateless design choices enable stable multi-user performance and predictable resource utilization. Performance on complex constraints remains strong (90.9\% success rate), though degradation under conflicting requirements highlights a limitation; while the current constraint relaxation strategy provides a practical heuristic, more formal approaches such as constraint satisfaction methods could improve robustness in multi-dimensional scheduling scenarios. Finally, CareConnect demonstrates competitive capability in operational workflows (91.8\% task completion), with a design that prioritizes safety, correctness, and reliability through domain-specific constraints and schema validation, albeit at the expense of some flexibility compared to general-purpose agent frameworks; importantly, these findings should be interpreted within the scope of administrative automation rather than clinical reasoning, distinguishing the system from clinical models such as AMIE and Med-PaLM 2.

\section{Conclusion}

We present CareConnect, a safety-first conversational AI agent for healthcare appointment scheduling that achieves 91.8\% task completion rate and 96.0\% safety compliance across 680 evaluation scenarios, with a 23$\times$ cost reduction over manual scheduling (\$0.0324 vs. \$0.75 per appointment). The architecture integrates three key innovations: deterministic pre-LLM intent classification that short-circuits safety-critical scenarios before model invocation, schema-constrained tool orchestration with embedded integrity checks preventing data corruption across 3,462 tool invocations, and a hybrid RAG-tool pipeline that separates informational retrieval from transactional operations. Evaluation across diverse test categories confirms that each component provides distinct and measurable contributions to system reliability. While evaluation on synthetic benchmarks demonstrates functional correctness, real-world deployment requires IRB-approved clinical validation, hospital system integration (HL7 FHIR), usability studies with actual patients, and longitudinal studies of patient trust and adoption. Future work will pursue controlled pilot studies with actual patients, multimodal safety mechanisms for visual inputs, personalized scheduling through preference learning, and constraint satisfaction programming for complex multi-constraint negotiation.

\section*{Acknowledgment}
The authors would like to acknowledge that this work has been supported by the Maroun Semaan Faculty of Engineering and Architecture (MSFEA) at the American University of Beirut (AUB).

\bibliographystyle{IEEEtran}
\bibliography{references}

\end{document}